\pdfoutput=1

\documentclass[11pt]{article}

\usepackage{ACL2023}

\usepackage{times}
\usepackage{latexsym}

\usepackage[T1]{fontenc}

\usepackage[utf8]{inputenc}

\usepackage{microtype}

\usepackage{inconsolata}

\usepackage{multirow}
\usepackage{makecell}

\usepackage{times}
\usepackage{latexsym}
\usepackage{graphicx}
\usepackage{subfigure}
\usepackage{float}
\graphicspath{{images}}
\usepackage{stfloats}
\usepackage{booktabs}
\usepackage{indentfirst}
\usepackage{amsmath}
\usepackage{amssymb}

\usepackage{enumitem}

\usepackage{colortbl}
\usepackage{arydshln}

\usepackage[linesnumbered,ruled,vlined]{algorithm2e}

\SetCommentSty{mycommfont}

\SetKwInput{KwInput}{Input}                
\SetKwInput{KwOutput}{Output}              

\usepackage{CJKutf8}

\usepackage{listings}
\usepackage{xcolor}

\definecolor{codegreen}{rgb}{0,0.6,0}
\definecolor{codegray}{rgb}{0.5,0.5,0.5}
\definecolor{codepurple}{rgb}{0.58,0,0.82}
\definecolor{backcolour}{rgb}{0.95,0.95,0.92}

\lstdefinestyle{mystyle}{
    backgroundcolor=\color{backcolour},   
    commentstyle=\color{codegreen},
    keywordstyle=\color{magenta},
    numberstyle=\tiny\color{codegray},
    stringstyle=\color{codepurple},
    basicstyle=\ttfamily\small,
    breakatwhitespace=false,         
    breaklines=true,                 
    captionpos=b,                    
    keepspaces=true,                 
    numbers=left,                    
    numbersep=5pt,                  
    showspaces=false,                
    showstringspaces=false,
    showtabs=false,                  
    tabsize=2
}

\title{PARA: Parameter-Efficient Fine-tuning with \underline{P}rompt \underline{A}ware \underline{R}epresentation \underline{A}djustment}

\author{
Zequan Liu$^1$$^{*}$ \ \ 
Yi Zhao$^2$\thanks{\ \ Equal contributions. } \ \
Ming Tan$^3$ \ \ 
Wei Zhu$^4$\thanks{\ \ Corresponding author. Email: michaelwzhu91@gmail.com. }  \ \ 
Aaron Xuxiang Tian $^5$  \\ 
\textsuperscript{\rm 1} RWTH Aachen University, Aachen, Germany  \\
\textsuperscript{\rm 2} University of Pennsylvania, USA \\
\textsuperscript{\rm 3} Southern University of Science and Technology, Shenzhen, China\\
\textsuperscript{\rm 4} University of Hong Kong, Hong Kong, China  \\
\textsuperscript{\rm 5} Carnegie Mellon University, Pittsburgh, USA \\
}

\begin{document}
\maketitle
\begin{abstract}

In the realm of parameter-efficient fine-tuning (PEFT) methods, while options like LoRA are available, there is a persistent demand in the industry for a PEFT approach that excels in both efficiency and performance within the context of single-backbone multi-tenant applications. This paper introduces a new and straightforward PEFT technique, termed \underline{P}rompt \underline{A}ware \underline{R}epresentation \underline{A}djustment (PARA). The core of our proposal is to integrate a lightweight vector generator within each Transformer layer. This generator produces vectors that are responsive to input prompts, thereby adjusting the hidden representations accordingly. Our extensive experimentation across diverse tasks has yielded promising results. Firstly, the PARA method has been shown to surpass current PEFT benchmarks in terms of performance, despite having a similar number of adjustable parameters. Secondly, it has proven to be more efficient than LoRA in the single-backbone multi-tenant scenario, highlighting its significant potential for industrial adoption.


\end{abstract}

\begin{CJK*}{UTF8}{gbsn}

\section{Introduction}

large language models (LLMs) draw attention by achieving amazing performance on a wide range of tasks \cite{yue2024tcmbench,qin2023chatgpt,PromptCBLUE,text2dt_shared_task,Text2dt,zhu_etal_2021_paht,Li2023UnifiedDR,Zhu2023BADGESU,Zhang2023LECOIE,Zhu2023OverviewOT,guo-etal-2021-global,zhu-etal-2021-discovering,Zheng2023CandidateSF,info:doi/10.2196/17653,Zhang2023NAGNERAU,Zhang2023FastNERSU,Wang2023MultitaskEL,Zhu2019TheDS,Zhu2021LeeBERTLE,Zhang2021AutomaticSN,Wang2020MiningIH}. In industrial applications, LLMs are frequently utilized in a single-instance, multi-tenant configuration, as highlighted in Chen et al.'s 2023 study on PunicaML \cite{Chen2023PunicaML}. An instance of this is when an LLM vendor offers a model as a service (MaaS), as described by \citet{gan2023model}. In this arrangement, various clients can tailor the LLM to their specific needs using their own parameter-efficient fine-tuning (PEFT) modules. A locally installed LLM is typically required to manage a variety of tasks for different tenants, each with their own set of PEFT parameters. However, while techniques like Low-Rank Adaptation (LoRA) \cite{hu2021lora} are adept at fine-tuning LLMs, they add considerable latency to each generation step because the low-rank components cannot be integrated into the main model structure. On the other hand, (IA)$^{3}$ \cite{Liu2022FewShotPF}, which relies solely on dot product operations, is a more efficient PEFT approach but may lack the necessary expressiveness. Consequently, there is a pressing need in the industry for a PEFT method that strikes a balance between efficiency and effectiveness.

\begin{figure*}
\centering
\includegraphics[width=0.64\textwidth]{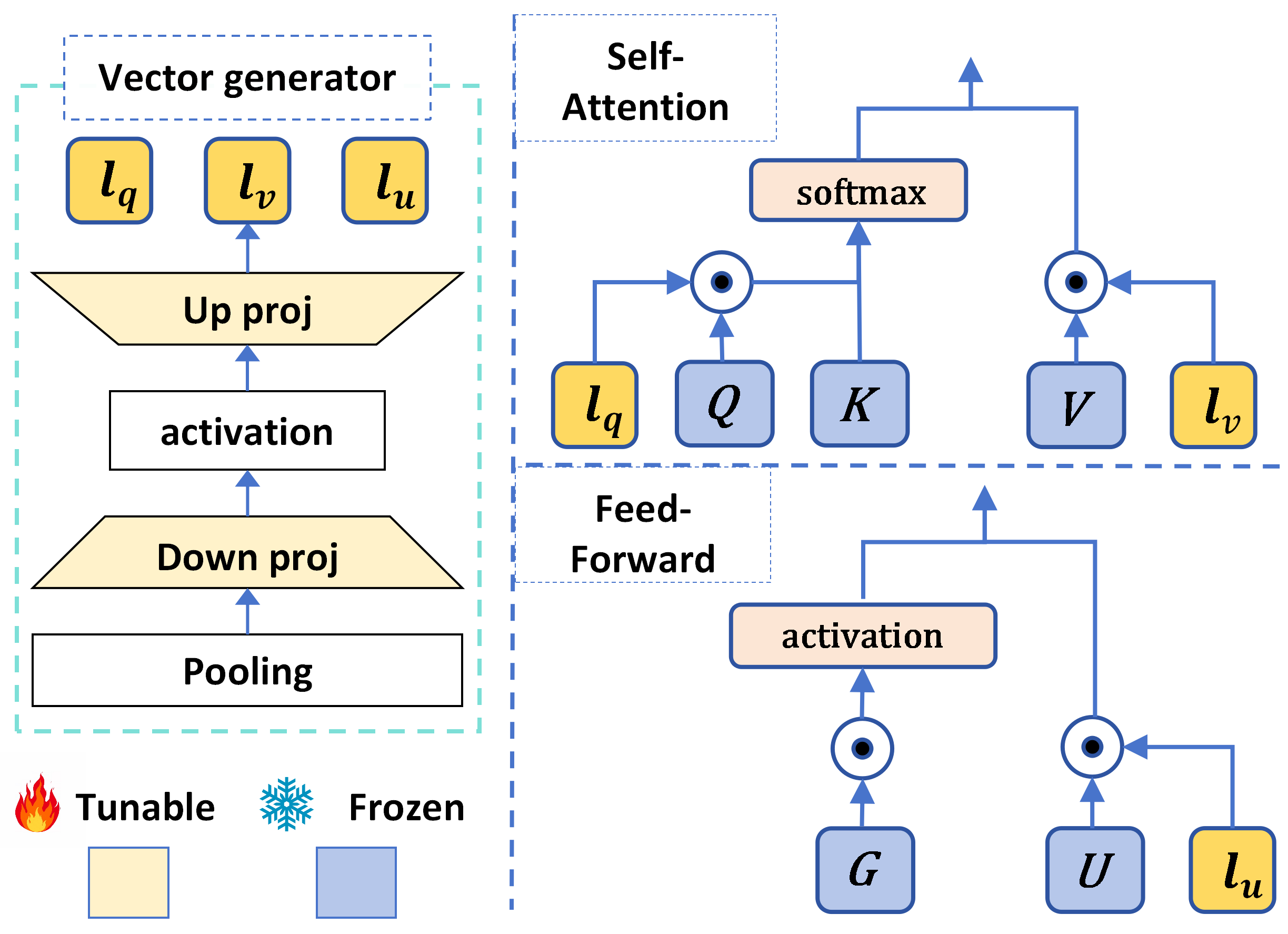}
\caption{A schematic representation of our PARA approach is depicted below. On the left, the vector generator is composed of several components, including a pooler, a down-projection layer, an activation function, and an up-projection layer. This generator takes the hidden states of the prompt as input and produces adjusting vectors as output. On the right, these adjusting vectors are used to scale the Query (Q) and Value (V) hidden states within the MHSA (Multi-Head Self-Attention) module, as well as the Up (U) hidden states within the feed-forward network. }
\label{fig:architecture}
\end{figure*}

In this work, we propose a novel PEFT method called \underline{P}rompt \underline{A}ware \underline{R}epresentation \underline{A}djustment (PARA) (depicted in Figure \ref{fig:architecture}). Our method fine-tuned the LLMs by directly modifying the hidden representations in the model by multiplying them by adjusting vectors and thus regulating the behaviors of LLMs. Unlike the previous literature like \citet{Liu2022FewShotPF} or \citet{BenZaken2021BitFitSP}, we introduce a novel prompt-aware mechanism to the PEFT method. The adjusting vectors are not randomly initialized and fixed across different input prompts. Instead, we install a vector generator (VG) before each Transformer layer, taking the input prompts' hidden states as input and generating the adjusting vectors as outputs. VG is a lightweight bottleneck architecture consisting of a pooling layer, a down-projection layer, an activation function, and an up-projection layer.

We perform a wide range of experiments across a diverse set of tasks to establish the efficacy of our PARA approach. It's important to note that our approach consistently surpasses robust PEFT benchmarks with similar adjustable parameter limits, particularly the latest LoRA iterations, (IA)$^3$ \cite{Liu2022FewShotPF}, and BitFit \cite{BenZaken2021BitFitSP}. We also demonstrate that our method exhibits substantially reduced latency in a multi-tenant environment compared to LoRA-based approaches, highlighting its suitability for real-world applications.

Our contributions can be encapsulated as follows:
\begin{itemize}
\item We introduce an innovative PEFT technique, PARA, which refines LLMs by producing adjustment vectors based on input prompts to alter the hidden states of LLMs.

\item Our comprehensive experiments and analyses reveal that our PARA system is (a) feasible and surpasses the competition within comparable parameter constraints. (b) swift during the inference phase for LLMs.

\end{itemize}

\section{Related works}

Parameter-efficient fine-tuning (PEFT) entails selectively optimizing a subset of parameters within a large pre-trained model while leaving the core model architecture intact for adaptation purposes \cite{liu2024alora,tian2024fanlora,zheng2024sca,zhang2024milora,Ding2022DeltaTA,Zhang2023LearnedAA,zhu-tan-2023-spt,Cui2023UltraFeedbackBL,zheng2024nat4at,zhu2023acf,gao2023f,zuo-etal-2022-continually,zhang-etal-2022-pcee,sun-etal-2022-simple,zhu-etal-2021-gaml,Zhu2021MVPBERTMP,li-etal-2019-pingan,zhu2019panlp,zhu2019dr,zhou2019analysis}. In contrast, addition-based techniques involve integrating extra neural components or parameters into the existing model framework. Notable contributions in this domain include Adapter \cite{houlsby2019parameter,Rckl2020AdapterDropOT,Zhang2023LearnedAA}, Prefix tuning \cite{Li2021PrefixTuningOC}, Prompt tuning \cite{lester2021power}, P-tuning V2 \cite{Liu2022PTuningPT,zhu2024iapt}, (IA)$^{3}$ \cite{Liu2022FewShotPF}, and BitFit \cite{BenZaken2021BitFitSP}. Conversely, specification-based methods involve the explicit designation of parameters that are either adjustable or subject to pruning \cite{BenZaken2021BitFitSP,guo-etal-2021-parameter,zhao-etal-2020-masking,zheng2024chimera}. The reparameterization-based strategies have garnered significant interest \cite{hu2021lora}. These approaches convert the parameters being optimized into a format that is both low-rank and parameter-efficient. Such PEFT methods are underpinned by the insight that the dimensionality intrinsic to fine-tuning is relatively low \cite{aghajanyan-etal-2021-intrinsic}. LoRA \cite{hu2021lora}, for instance, posits that the variation in weights during tuning is characterized by a low intrinsic rank, and thus focuses on optimizing the low-rank factorization of the weight matrix changes. PEFT techniques have found broad application, particularly with the rise of open-source large-scale language models \cite{2023arXiv230318223Z} and the trend of tailoring these models to specific use cases through instruction tuning \cite{alpaca,2023arXiv230514314D}.

In this research, we introduce a novel framework known as PARA, which is designed for the parameter-efficient fine-tuning of Large Language Models (LLMs). This approach not only enhances efficiency during LLM inference but also delivers superior performance across various downstream applications.

\section{Methods}

\subsection{Preliminaries}

\noindent \textbf{Transformer model} \quad Currently, the most widely used open-sourced large language models adopt the stacked Transformer architecture \cite{Vaswani2017AttentionIA}. The transformer block is primarily constructed using two key submodules: a multi-head self-attention (MHA) layer and a fully connected feed-forward (FFN) layer. Denote the input sequence's length as $l$, the hidden states' dimension as $d_{model}$, and the dimension at the FFN module as $d_{ffn}$. The MHA is given as follows:\footnote{We omit the multi-head setting for simplicity of illustrations. }
\begin{equation}
\text{softmax} \left( \dfrac{ Q K }{\sqrt{d_{model}} } \right) V,
\label{eq:self_attn_1}
\end{equation}
where $Q = xW^{Q}$, $K = xW^{K}$, $V = xW^{V}$, $x \in \mathbf{R}^{l\times d_{model} }$ is the input tensor. $W^{Q}, W^{K}, W^{V} \in \mathbf{R}^{d_{model} \times d_{model} }$ are the query, key, and value projection layers (denoted as the Query, Key, and Value modules, or the Q, K, V modules). The FFN module consists of linear transformations and an activation function $g^{ffn}$ such as ReLU or GELU \cite{Hendrycks2016GaussianEL}. Take the FFN module in the LlaMA-2 models \cite{Touvron2023Llama2O} as example:
\begin{equation}
(g^{ffn}( G ) * U ) W^{D}, 
\label{eq:ffn_1}
\end{equation}
where $G = xW^{G}$, $U = xW^{U}$, $W^{G}, W^{U} \in \mathbf{R}^{ d_{model} \times d_{ffn} } $ (denoted as Gate and Up module, or the G and U modules).

\noindent \textbf{Task formulation} \quad Denote the task's training set as $\mathcal{D}_{\text{train}} = {(x_m, y_m), m = 1, 2, ..., M}$, where $M$ represents the number of samples. In this work, we only consider the case where input $x_m$ and target $y_m$ are both text sequences. 


\subsection{PARA}

Now we present the framework of our novel \underline{P}rompt \underline{A}ware \underline{R}epresentation \underline{A}djustment (PARA) method.

\noindent \textbf{Formulation} \quad Denote the hidden state of the input prompt with length $T_{ins}$ at the current Transformer layer as $\mathbf{h}$. As shown in Figure \ref{fig:architecture}, the vector generator $\text{VG}()$ use $\mathbf{h}$ as input to generate three learned vectors, $l_{q}, l_{v} \in \mathbb{R}^{d_{model} }$ and $ l_{u} \in \mathbb{R}^{d_{ffn} }$, with a vector generator:
\begin{equation}
l_{q}, \ l_{v}, \ l_{u} = \text{VG}( \mathbf{h} ),
\label{eq:vector_generator}
\end{equation}
and these generated vectors are used to modify the hidden representations in the self-attention and FFN modules. Thus, under PARA, the self-attention mechanism of Transformer (in Equation \ref{eq:self_attn_1}) is changed to 
\begin{equation}
\text{softmax} \left(  Q^{'} K / \sqrt{d_{model}} \right) V^{'},
\label{eq:self_attn_2}
\end{equation}
where $Q^{'} = l_{q} \odot Q$, $V^{'} = l_{v} \odot V$, and $\odot$ denotes the element-wise dot product. And the FFN module (Equation \ref{eq:ffn_1}) is modified to 
\begin{equation}
(g^{ffn}( G ) \odot U^{'} )W^{D},
\label{eq:ffn_2}
\end{equation}
where $U^{'} = l_{u} \odot U$.\footnote{We use the "broadcasting notation" in the Equations \ref{eq:self_attn_2} and \ref{eq:ffn_2}. Take so that the $(m, n)$-th entry of $U^{'}$ is $l_{u}[n] \odot U[m, n]$. }\footnote{From our preliminary experiments, we find that generating adjustment vectors for the other hidden states like $\text{K}$ and $\text{G}$ will not result in clear performance gains.}

\noindent \textbf{Vector generator} \quad Now we introduce the central component of our PARA framework, the vector generator denoted as $\text{VG}()$. This function accepts $\textbf{h}$ as its input and is composed of a pooling module along with a pair of projection operations, each accompanied by an activation function. The process begins by converting $\textbf{h}$ into a single vector using the $\text{Pooler}()$ function. In line with the works of \citet{radford2018improving} and \citet{lewis2019bart}, $\text{Pooler}()$ outputs the vector representation corresponding to the final token in the prompt. Subsequently, the pooled vector is projected from the dimension $d_{model}$ down to $r < d_{model}$ through a projection layer defined by $W^{vg}_{down} \in \mathbb{R}^{d_{model} \times r}$. This is followed by the application of an activation function $g^{vg}$, after which the vector is projected to the dimension $d_{out} = 2 * d_{model} + d_{ffn}$ via another projection layer, utilizing the weight matrix $W^{vg}_{up}$ and bias term $b_{up}^{vg}$. Mathematically, the vector generator can be expressed by the following equations:
\begin{align}
 & l = (g^{vg}(\text{Pooler}(\textbf{h}) W^{vg}_{down} )) W^{vg}_{up} + b_{up}^{vg}, \nonumber \\
 & l_{q}, \ l_{v}, \ l_{u} = \text{Split}( l ), 
\end{align}
where the $\text{Split}()$ function is responsible for dividing the vector into three separate vectors, each of dimension $d_{model}$, $d_{model}$, and $d_{ffn}$, respectively.

The concept of prompt-awareness is derived from studies on in-context learning. As shown by \citet{rubin2022learning} and \citet{Li2023UnifiedDR}, enhancing the performance of Large Language Models (LLMs) can be achieved by dynamically creating an expanded prompt that includes examples tailored to the specific input prompt. It has been observed that distinct input prompts necessitate unique examples to evoke more effective responses from LLMs. Similarly, the idea of tailoring PEFT parameters to the input prompt could enhance the method's expressive capabilities and more precisely control the conduct of LLMs.

It's important to recognize that causal language models (CLM), which are based on decoders, often utilize the KV cache mechanism\footnote{\url{https://www.dipkumar.dev/becoming-the-unbeatable/posts/gpt-kvcache/}} to enhance efficiency during the generation process. The vector generators in our system integrate flawlessly with this KV cache mechanism. This is because the vectors $l_{q}$, $l_{v}$, and $l_{u}$ are produced when the input instruction (or prompt) is initially processed by the LLM. These vectors are then reused for the generation of subsequent tokens, and the vector generators are not invoked again. On the other hand, the LoRA method introduces reparameterizations to the model's parameters, necessitating that its low-rank weight matrices be included in the forward calculations for each token generation step, which results in increased latency.

\section{Experiments}

In this section, we conduct experiments to evaluate our PARA method.

\subsection{Baselines}

We compare our PARA framework with the current SOTA PEFT baseline methods: (a) (IA)$^{3}$ \cite{Liu2022FewShotPF}, which multiplies learnable vectors to the hidden representations of LLMs. (b) Houlsby-Adapter \cite{houlsby2019parameter}. (c) Learned-Adapter \cite{Zhang2023LearnedAA}. (d) LoRA \cite{hu2021lora}. (e) AdaLoRA \cite{Zhang2023AdaptiveBA}. (f) SSP \cite{Hu2022SparseSS}, which combines different PEFT methods. The baselines are implemented using Transformers \cite{wolf2020transformers} or their open-sourced codes.

\subsection{Datasets and evaluation metrics}

We experiment on the following benchmark tasks: (a) three benchmark question-answering tasks: SQuAD \cite{rajpurkar-etal-2016-squad} and two tasks from the SuperGLUE benchmark~\cite{Wang2019SuperGLUEAS} (BoolQ, COPA). (b) two widely used LLM evaluation benchmarks, MT-Bench \cite{2023arXiv230605685Z}, MMLU \cite{hendrycks2020measuring}. (c) A proprietary LLM evaluation benchmark, LLM-Eval1, for internal LLM developments of an industrial participant. (d) a proprietary high-school-level mathematical solving dataset, HSM10K. (e) a proprietary SQL generation task, Q2SQL. The above tasks' dataset introductions, statistics, and evaluation metrics are detailed in Appendix \ref{sec:appendix_datasets}.

\begin{table*}[tb!]
\centering
\resizebox{1.0\textwidth}{!}{
\begin{tabular}{cccccccc}
\hline
Datasets  &  \#train    &  \#dev   &   \#test   &   $ | \mathcal{Y} | $   &   Type   &  Labels  &  Metrics  \\ 
\hline
BoolQ  &  9.4k   &    1.6k   &  1.6k   &    2   &   Question Answering    &  True, False   &   acc \\
COPA   &   0.4k   &   0.05k   &   0.05k    &   2   &      Question Answering   &    choice1, choice2   &   acc   \\
SQuAD &   87k  &  1k    &   5.9k   &   -   &  Question Answering  &  -   &   f1-em   \\

\hdashline

MT-Bench   &   -   &  -  &   80  &   -   &    Question Answering   &  -    &  GPT-4 scores       \\

MMLU  &  -  &  1.5k  &  14.1k   &    -   &   Question Answering    &   -    &  acc    \\

\hdashline

HSM10K  &   9K  &  0.6K  &  0.7K   &  - &   Math reasoning  &   -    &   acc  \\
Q2SQL  &   60k  &  4K  &  10K   &   -  &   SQL generation  &   -    &   acc  \\

\hdashline

LLM-Eval1  &  -  &  -  &  3.6k   &     -   &    Question Answering    &   -    &  acc    \\

UltraChat   &    766k  &    7.7k    &  -   &  -  &  Instruction tuning  &  -   &   -   \\

\hline
\end{tabular}}
\caption{\label{tab:dataset_stats} The statistics of the datasets evaluated in this work. $ | \mathcal{Y} | $ is the number of classes for a classification task. }
\end{table*}

\begin{table*}[tb!]
\centering
\resizebox{0.88\textwidth}{!}{
\begin{tabular}{c|c|ccccc}
\hline
\multirow{2}*{\textbf{Method}}   &   \textbf{Tunable}   &    \textbf{HSM10K}    &    \textbf{Q2SQL}   &     \textbf{SQuAD}    &   \textbf{BoolQ}  &  \textbf{COPA}       \\ 

&  \textbf{Params}  &   \textbf{(acc)}   &   \textbf{(acc)}    &   \textbf{(f1-em)}    &  \textbf{(acc)}   &   \textbf{(acc)}     \\
\hline

Full-FT  & 7B   &     57.9   &  82.9  &    89.5    &     88.7   &   91.9    \\
\hline

\multicolumn{7}{c}{\textbf{\emph{Baselines PEFT methods}}}  \\
\hline

Housbly-Adapter   &    9.4M    &    52.8   &   80.4  &   87.3  &   84.5   &     90.4    \\ 
Learned-Adapter   &   9.5M   &     53.7  &  81.3  &    87.6   &  85.9     &     90.6     \\
\hdashline

SSP &   8.6M   &  54.6  &  81.5  &     87.4    &      86.4       & 
   91.1  \\
(IA)$^{3}$  &    9.8M   &    54.3    &  81.2  &      87.6     &    86.2   &     90.7   \\

\hdashline
LoRA   &     10.0M   &     55.1  &    81.8   &       \underline{87.7}    &    86.3      &    90.9   \\
AdaLoRA   &  10.0M   &   \underline{55.6}  &   \underline{82.2}  &       87.5    &     \underline{87.0}   &    \underline{91.2}    \\ 

\hline
\multicolumn{7}{c}{\textbf{\emph{Our proposed method}}}  \\
\hline

PARA   &  8.9M    &     \textbf{56.3}  &    \textbf{82.8}    &     \textbf{88.5}    &    \textbf{87.7}    &   \textbf{92.0}   \\

\hline
\end{tabular}}

\caption{\label{tab:results_main_1} The Overall comparison of the SQuAD, BoolQ, COPA, HSM10K and Q2SQL tasks. The backbone model is LLM-Assist 7B. We report the median performance over five random seeds. Bold and Underline indicate the best and the second-best results. The metric for each task is explained in Appendix \ref{sec:appendix_evaluations}. } 
\end{table*}

\subsection{Experiment Settings}

\noindent\textbf{Computing infrastures} \quad We run all our experiments on NVIDIA A40 (48GB) GPUs. 

\noindent\textbf{Pretrained backbones} \quad The main experiments use the most recent open-sourced LLM, LlaMA-2 7B released by Meta \cite{Touvron2023Llama2O} as the pretrained backbone model. We will also use the LlaMA-2 13B model and Gemma 2B \cite{team2024gemma} in the ablation studies.

\noindent\textbf{Prediction heads} \quad After receiving a prompt or instruction, all the predictions are generated using the language modeling head (LM head). For decoding during inference, we use beam search with beam size 3.











\noindent\textbf{Hyper-parameters for the PARA framework} \quad In our experiments, unless otherwise specified, we set: (a) the bottleneck dimension $r$ of the PARA vector generator to 12, (b) the activation function $g^{vg}$ to the GeLU activation function \cite{Hendrycks2016GaussianEL}. (c) The $W_{down}^{vg}$ is initialized with a Gaussian distribution of mean 0 and std 0.02. $W_{up}^{vg}$ is zero initialized, and $b_{up}^{vg}$ is initialized with ones. Under the above settings, our PARA method will introduce 8.9M tunable parameters to LlaMA-2 7B.

\noindent\textbf{Training settings for PARA} \quad Utilizing the HugginFace Transformers \cite{wolf-etal-2020-transformers}, PEFT \cite{peft}, or the original code repositories, we implement all the methods for training and prediction tasks. When fine-tuning the LlaMA-2 7B model, the sequence length is capped at 2048. The training epochs are limited to a maximum of 10. The batch size is adjusted to 16 for tasks with fewer than 10k training samples, and 128 for larger datasets. AdamW serves as the optimizer, employing a linear learning rate decay strategy with a 6\% warm-up period over the training steps. The learning rate is configured at 1e-4. All other hyper-parameters align with those used by \citet{wolf-etal-2020-transformers}. The model's performance is assessed on the development set every 200 steps. Early stopping is initiated with a patience level of 10, meaning training will be halted if the model fails to record a lower loss on the development set for 10 consecutive evaluations. The optimal checkpoint identified on the development set is then applied to make predictions on the test set.

\noindent\textbf{Reproducibility} \quad We run each task under five different random seeds and report the median performance on the test set of each task.

\subsection{Main results}
\label{subsec:main_results}

The outcomes of our experiments on the SQuAD, BoolQ, COPA, HSM10K, and Q2SQL benchmarks are detailed in Table \ref{tab:results_main_1}, where the count of adjustable parameters is listed in the second column. The data in Table \ref{tab:results_main_1} indicates that our PARA approach surpasses the standard methods on all five benchmarks, with an equivalent or reduced number of adjustable parameters. Notably, PARA achieves better results than the previous state-of-the-art LoRA-style baselines, namely LoRA and AdaLoRA, while using a similar parameter count.

\begin{table*}[tb!]
\centering
\resizebox{0.66\textwidth}{!}{
\renewcommand\arraystretch{1.05}
\begin{tabular}{c|ccc}
\hline
\multirow{2}*{\textbf{Method} }  &    \textbf{MT-Bench}     &    \textbf{MMLU}    &   \textbf{LLM-Eval1}   \\ 

&   \textbf{gpt4-score} ($\uparrow$)   &    \textbf{acc}   &    \textbf{acc}    \\ 
\hline
AdaLoRA    &   7.13   &    46.5    &   56.8   \\
\hdashline
PARA   &   7.21  &  47.4  &   57.7     \\
\hline

\end{tabular}}
\caption{\label{tab:results_llm_eval} Performance of general-purpose instruction tuning using the PARA and AdaLoRA methods. The backbone model is LLM-Assist 7B. $\uparrow$ means the metric is higher the better.}
\end{table*}

After fine-tuning the LLM-Assist 7B model on the UltraChat dataset \cite{ding2023enhancing} using our PARA configuration or the AdaLoRA techniques, we proceed to assess its performance on the demanding benchmarks: MT-Bench, M
MLU, and LLM-Eval1. The trials are executed in a zero-shot scenario, with no exemplar instances appended to the input prompts. The outcomes are detailed in Table \ref{tab:results_llm_eval}. Aligning with the findings from the prior experiments (Table \ref{tab:results_main_1}), our PARA approach surpasses the AdaLoRA techniques across the three benchmarks, indicating that PARA is more effective in bolstering the directive tuning proficiency of expansive language models.

\begin{table*}[tb!]
\centering
\resizebox{0.7\textwidth}{!}{
\begin{tabular}{c|ccc}
\hline
\textbf{Method}   &    \textbf{Beam size}  &  \textbf{Speed (tps)}   &   \textbf{Memory cost (MiB)}     \\ 
\hline

\multirow{ 2}{*}{ LoRA }   &   1    &   25.1    &   14616   \\
    &   3   &    21.9    &    16104  \\

\hdashline
\multirow{ 2}{*}{ (IA)$^{3}$ }   &   1    &  33.1    &    14572    \\
&   3   &    27.6   &   16036    \\

\hdashline
\multirow{ 2}{*}{ PARA }   &   1    &  32.8    &     14512     \\
&   3   &    27.6     &     15986    \\
    
\hline
\end{tabular}}
\caption{\label{tab:results_efficiency_analysis} The memory and speed of LlaMA-2 7B for generating responses with different PEFT methods. }
\end{table*}

\subsection{Further analysis}

\noindent\textbf{Analysis of the inference efficiency} \quad To showcase the inference efficiency of our PARA approach, we proceed to juxtapose the GPU memory usage and the rate of generation for PARA, LoRA, and (IA)$^{3}$. In the course of this experiment, parameters of LoRA have not been integrated into the main model to emulate a single-LLM multi-tenant configuration as indicated in \cite{Chen2023PunicaML}. We have capped the creation of new tokens to 32, utilizing beam search with a beam width of either 1 or 3. The initial instruction's length is set at 274, employing the LlaMA-2 tokenizer. We execute the generation process a total of 100 instances to ascertain the average metric estimates, thereby diminishing the element of randomness. We introduce two key metrics for gauging efficiency: (a) the apex memory expenditure during the generation phase, and (b) the rate of token generation per second (tps). The comparative data is delineated in Table \ref{tab:results_efficiency_analysis}.

As depicted in Table \ref{tab:results_efficiency_analysis}, it is evident that: (a) our PARA approach possesses a similar number of adjustable parameters, memory usage, and generation rate to (IA)$^{3}$. (b) PARA outperforms LoRA in terms of speed. The enhanced velocity of PARA over LoRA can be attributed to several elements: (i) our vector generation process is both minimal and efficient during the inference phase. (ii) The vectors, $l_q$, $l_v$, $l_u$, are generated solely upon the input of instructions to the LLM and prior to the creation of the initial new token. These vectors are then reused in subsequent generation stages with the aid of KV-cache, eliminating the need for repeated invocation of the vector generators. Conversely, the LoRA technique necessitates the model to engage the LoRA modules at every generation stage, leading to increased latency.

\begin{table*}[tb!]
\centering
\resizebox{0.4\textwidth}{!}{
\begin{tabular}{c|cc}
\hline
\multirow{2}*{\textbf{Method}}    &     \textbf{BoolQ}     &       \textbf{SQuAD}  \\ 

&    \textbf{(acc)}  &    \textbf{(f1-em)}  \\
\hline 

\multicolumn{3}{c}{\textbf{\emph{Results for LlaMA-2 13B model }}}  \\
\hline

(IA)$^{3}$     &    89.6     &   90.6  \\
LoRA   &      90.0   & 90.9        \\
AdaLoRA   &    90.2   &   91.6     \\
\hdashline
PARA   &   \textbf{90.9}   &     \textbf{92.1}     \\

\hline 
\multicolumn{3}{c}{\textbf{\emph{Results for Gemma 2B }}}  \\
\hline

(IA)$^{3}$    &   82.7      &   78.1      \\
LoRA   &         82.8      &    78.4  \\
AdaLoRA   &      83.0      &   78.8   \\ 
\hdashline
PARA   &     \textbf{83.6}   &    \textbf{79.7}    \\

\hline
\end{tabular}}
\caption{\label{tab:results_different_backbones} Results for different PEFT methods on the BoolQ and SQuAD benchmarks. The backbone LMs are LlaMA-2 13B and Gemma 2B. The metrics are explained in Appendix \ref{sec:appendix_evaluations}.}
\end{table*}

\noindent\textbf{Ablation on the pretrained backbones} \quad Our principal experiments were carried out utilizing the LlaMA-2 7B model. In order to showcase the versatility of our approach, additional experiments have been executed on both the LlaMA-2 13B model and the Gemma 2B model. The corresponding outcomes are detailed within Table \ref{tab:results_different_backbones}. Furthermore, our approach surpasses the performance of the foundational methodologies on these alternative model architectures.

\section{Conclusion}

This study introduces PARA, an innovative approach for the parameter-efficient fine-tuning of expansive language models. We integrate a vector generator within each Transformer layer to produce adjustment vectors that modulate the functionality of the LLM core. The vector generator utilizes the hidden states of the input prompts as inputs and features a lightweight bottleneck design. PARA offers greater efficiency in inference compared to LoRA, as it operates harmoniously with the KV-cache system. Our experiments across a range of tasks show that PARA surpasses the performance of standard methods while maintaining high inference efficiency. PARA is advantageous for industrial applications that leverage LLMs.

\section*{Limitations}

We showed that our proposed method can greatly improve the performance of parameter-efficient tuning on diverse tasks and different pretrained models (i.e., LlaMA-2 7B, LlaMA-2 13B model and Gemma 2B), while maintaining efficiency during inference. However, we acknowledge the following limitations: (a) the more super-sized open-sourced LLMs, such as LlaMA-2 70B, are not experimented due to limited computation resources. (b) Other tasks in natural language processing, like information extraction, were also not experimented. But our framework can be easily transferred to other backbone architectures and different types of tasks. It would be of interest to investigate if the superiority of our method holds for other large-scaled backbone models and broader types of tasks. And we will explore it in future work.

\section*{Ethics Statement}

The finding and proposed method aims to improve the parameter-efficient tuning in terms of performance and efficiency. The used datasets are widely used in previous work and, to our knowledge, do not have any attached privacy or ethical issues. In this work, we have experimented with LlaMA-2, a modern large language model series. As with all LLMs, LlaMA-2’s potential outputs cannot be predicted in advance, and the model may in some instances produce inaccurate, biased or other objectionable responses to user prompts. However, this work's intent is to conduct research on different fine-tuning methods for LLMs, not building applications to general users. In the future, we would like to conduct further testing to see how our method affects the safety aspects of LLMs.

\bibliography{custom}
\bibliographystyle{acl_natbib}

\appendix

\section{Appendix for the datsets and evaluation metrics}
\label{sec:appendix_datasets}

\subsection{Datasets }

We now introduce the datasets we used for experiments. The detailed statistics of these tasks are presented in Table \ref{tab:dataset_stats}.

\noindent\textbf{COPA \& BoolQ} \quad These two tasks are question answering tasks in the format of binary choices, and are included in the SuperGLUE benchmark. Since the original test sets are not publicly available for these tasks, we follow \citet{Zhang2020RevisitingFB,Mahabadi2021CompacterEL} to divide the original validation set in half, using one half for validation and the other for testing.

\noindent\textbf{SQuAD task} \quad Stanford Question Answering Dataset (SQuAD) \cite{rajpurkar-etal-2016-squad} is a reading comprehension dataset, consisting of questions posed by crowdworkers on a set of Wikipedia articles, where the answer to every question is a segment of text, or span, from the corresponding reading passage, or the question might be unanswerable. This task is one of the most widely studied question answering task in the field. In this work, we use the v1.1 version of SQuAD. Since the original test sets are not publicly available for these tasks, we follow \citet{Zhang2020RevisitingFB,Mahabadi2021CompacterEL} and split 1k samples from the training set as the development set, and use the original development set as the test set. The detailed statistics of this task is presented in Table \ref{tab:dataset_stats}.

\noindent\textbf{HSM10K benchmark} \quad HSM10K is a dataset of 10.3K high quality high school level problems created by the math teachers. These problems are the most difficult ones from a wide source of math tests. The solving steps are generated by GPT-4 and then checked/rewritten by math teachers to ensure accuracy. We use this dataset to improve the math reasoning abilities of LLMs. The dataset is split into 9k/0.6K/0.7K train/dev/test sets.

\noindent\textbf{Q2SQL dataset} \quad Q2SQL consists of a corpus of 74K hand-annotated SQL query and natural language question pairs. This proprietary dataset is collected from a company in the health insurance company, where the SQL are primarily related to analyzing insurance policies. These SQL queries are further split into training (60k examples), development (4k examples) and test sets (10k examples). In this work, we will ask the LLMs to generate SQL queries based on the given natural language questions.

\noindent\textbf{The MMLU benchmark} \quad Massive Multitask Language Understanding (MMLU) \cite{hendrycks2020measuring} is a new benchmark designed to measure knowledge acquired during pretraining by evaluating large language models exclusively in zero-shot and few-shot settings. This makes the benchmark more challenging and more similar to how we evaluate humans. The benchmark covers 57 subjects across STEM, the humanities, the social sciences, and more. It ranges in difficulty from an elementary level to an advanced professional level, and it tests both world knowledge and problem solving ability. Subjects range from traditional areas, such as mathematics and history, to more specialized areas like law and ethics. The granularity and breadth of the subjects makes the benchmark ideal for identifying a model’s blind spots.

\noindent\textbf{MT-Bench} \quad The MT-Bench \cite{2023arXiv230605685Z} dataset is a widely used benchmark for evaluating the quality of LLMs. It contains 80 questions. The LLMs generate a two-round dialogue for these questions, and human annotators or LLM annotators will judge the quality of these responses.

\noindent\textbf{The LLM-Eval1 benchmark} \quad This benchmark is a proprietary dataset, designated to challenge the LLMs for reasoning, world knowledge, and task solving. This dataset is used internally to facilitate LLM development. LLM-Eval1 contains a suite of 47 challenging tasks from multiple domains including literature, healthcare, security, coding assistant, and software development and testing. The number of test samples are 3,569.

\noindent\textbf{The UltraChat dataset} \quad UltraChat \cite{ding2023enhancing} is an open-source, large-scale, and multi-round dialogue data curated with the help of OpenAI's GPT-3-Turbo API. To ensure generation quality, two separate GPT-3-Turbo APIs are adopted in generation, where one plays the role of the user to generate queries and the other generates the response. The user model is carefully prompted to mimic human user behavior and the two APIs are called iteratively to create a dialogue. There are 774k dialogues in the dataset, and we split it into a 99:1 train/validate set for the FanLoRA workflow.

\subsection{Evaluation metrics/protocols}
\label{sec:appendix_evaluations}

For the BoolQ and COPA tasks, we report accuracy following \cite{Wang2019SuperGLUEAS}.  

For the SQuAD dataset, we also report the average of the F1 score and the exact match score (denoted as f1-em).

For the HSM10K task, we will consider the correctness of the final answers. Thus, we report accuracy (denoted as acc). 

For the Q2SQL, we will consider the correctness of the generated SQL queries. A predicted SQL query is correct if and only if it can be executed and obtains the same results with the ground truth. 

For the MMLU and LLM-Eval1 tasks, we will directly consider the correctness of the final answers. Thus, we report accuracy (denoted as acc). 

For evaluating the quality of instruction tuned LLMs, we follow the practice of utilizing GPT-4 as a unbiased reviewer \cite{2023arXiv230605685Z}. 80 instructions from the MT-Bench is set as a test set. We generate model responses from a fine-tuned model with beam size 3 with the generation function in Huggingface Transformers \cite{wolf2020transformers}. Then we compare AdaLoRA and FanLoRA's answers with GPT-4. For each instruction in MT-Bench, GPT-4 \cite{gpt4} is asked to write a review for both answers from the two methods, and assigns a quantitative score on a scale of 10 to each response.

\end{CJK*}

\end{document}